\documentclass[letterpaper]{article} 
\usepackage{aaai2026}  
\usepackage{times}  
\usepackage{helvet}  
\usepackage{courier}  
\usepackage[hyphens]{url}  
\usepackage{graphicx} 
\usepackage{natbib}  
\usepackage{caption} 
\usepackage{algorithm}
\usepackage{algorithmic}

\usepackage{amsmath}
\usepackage{amsfonts}
\usepackage{subcaption}
\usepackage{pifont}
\usepackage{booktabs}
\usepackage{multirow}
\usepackage{colortbl}
\usepackage{xcolor} 

\usepackage{newfloat}
\usepackage{listings}
\DeclareCaptionStyle{ruled}{labelfont=normalfont,labelsep=colon,strut=off} 
\floatstyle{ruled}
\newfloat{listing}{tb}{lst}{}
\floatname{listing}{Listing}
\title{Explicit Temporal-Semantic Modeling for Dense Video Captioning via \\ Context-Aware Cross-Modal Interaction}
\author{
    Mingda Jia\textsuperscript{\rm 1,2}, Weiliang Meng\textsuperscript{\rm 1,2,*}, 
    Zenghuang Fu\textsuperscript{\rm 1,2}, 
    Yiheng Li\textsuperscript{\rm 1,2}, 
    Qi Zeng\textsuperscript{\rm 1,2},  \\
    Yifan Zhang\textsuperscript{\rm 1,2}, 
    Ju Xin\textsuperscript{\rm 3}, 
    Rongtao Xu\textsuperscript{\rm 4}, 
    Jiguang Zhang\textsuperscript{\rm 1,2,\#}, 
    Xiaopeng Zhang\textsuperscript{\rm 1,2}
}
\affiliations{

    \textsuperscript{\rm 1} MAIS, Institute of Automation, Chinese Academy of Sciences, Beijing, China\\
    \textsuperscript{\rm 2} School of Artificial Intelligence, University of Chinese Academy of Sciences\\
    \textsuperscript{\rm 3} Northern Navigation Service Center, Qingdao, China \\
    \textsuperscript{\rm 4} Spatialtemporal AI \\
    
    Corresponding author\{$^*,^\#$\}, email: $^*$weiliang.meng@ia.ac.cn;$^\#$jiguang.zhang@ia.ac.cn
}

\usepackage{bibentry}

\begin{document}

\maketitle

\begin{abstract}
Dense video captioning jointly localizes and captions salient events in untrimmed videos. Recent methods primarily focus on leveraging additional prior knowledge and advanced multi-task architectures to achieve competitive performance.
However, these pipelines rely on implicit modeling that uses frame-level or fragmented video features, failing to capture the temporal coherence across event sequences and comprehensive semantics within visual contexts.
To address this, we propose an explicit temporal-semantic modeling framework called Context-Aware Cross-Modal Interaction (CACMI), which leverages both latent temporal characteristics within videos and linguistic semantics from text corpus.
Specifically, our model consists of two core components: Cross-modal Frame Aggregation aggregates relevant frames to extract temporally coherent, event-aligned textual features through cross-modal retrieval; and Context-aware Feature Enhancement utilizes query-guided attention to integrate visual dynamics with pseudo-event semantics.
Extensive experiments on the ActivityNet Captions and YouCook2 datasets demonstrate that CACMI achieves the state-of-the-art performance on dense video captioning task.
\end{abstract}



\section{Introduction}

In recent years, video understanding has emerged as a rapidly growing focus within the fields of computer vision and multimodal analysis~\cite{wang2023internvid, song2024moviechat, nie2024slowfocus, chen2024sharegpt4video, li2024videomamba, wang2024internvideo2}, with video captioning recognized as a foundational task. Traditional video captioning aims to generate a concise description that summarizes the main content of a video, and significant progress has been achieved in this area~\cite{gao2017video, krishna2017dense, pei2019memory, seo2022end}. However, conventional methods often miss fine-grained details and struggle to handle multiple events or segments effectively. To overcome these limitations, dense video captioning (DVC) has been introduced, which seeks to produce descriptive annotations for all salient events in an untrimmed video, along with their precise temporal boundaries.

\begin{figure}[h]
  \centering
  \includegraphics[width=0.47\textwidth]{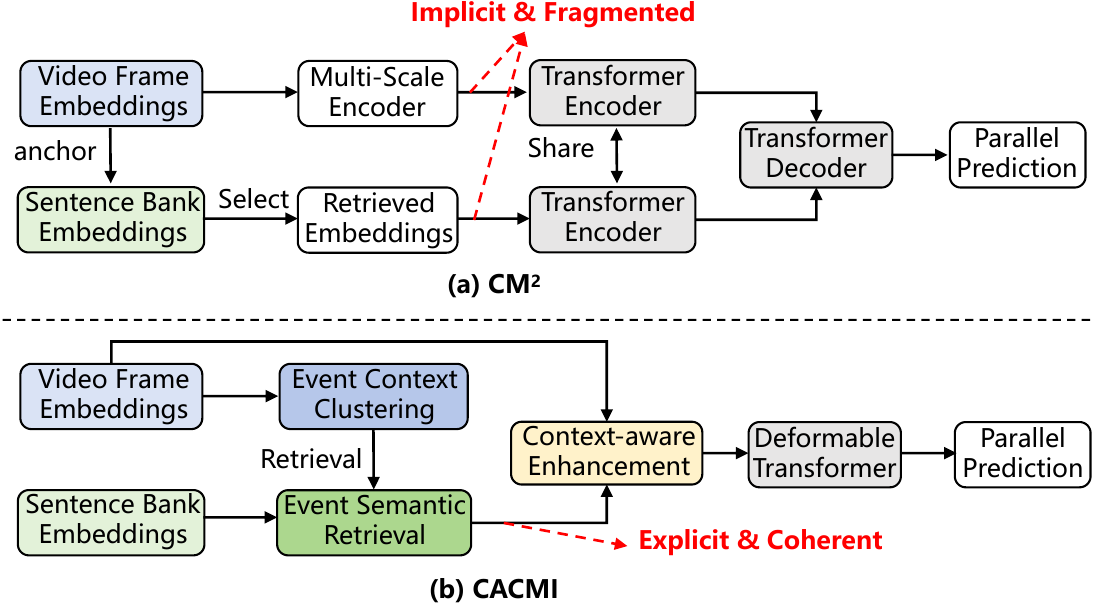}
  \caption{(a) CM$^{2}$ introduces a cross-modal memory-based model, the external sentence bank is specifically designed to select relevant implicit semantics.
  (b) Our CACMI harnesses explicit temporal-semantic information through context-aware cross-modal interaction to enhance the event localization and captioning performance. }
    \label{fig:fig1}
\end{figure}

Dense video captioning methods typically begin by leveraging a pre-trained image encoder to extract visual features from input frames, followed by the detection of salient event boundaries within these features to achieve temporal localization and event representations~\cite{zhou2018end, mun2019streamlined, wang2021end}. With the rapid advancement of vision-language models, recent approaches have explored retrieval-augmented generation for video captioning, incorporating external semantic knowledge into the encoding-decoding pipeline to enhance understanding and generation capabilities. In dense video captioning, the representative work CM$^{2}$~\cite{kim2024you} pioneered the integration of memory retrieval mechanism, effectively utilizing semantic cues from external sources to improve both event localization and caption generation.

Despite these advancements, recent memory-based methods depend on inherently implicit retrieval-augmented generation (RAG) frameworks~\cite{chen2023retrieval, ramos2023smallcap, kim2024you, kim2025hicm2}. These approaches employ manually designed windows for cross-modal retrieval at fragmented video segments, leading to two fundamental limitations:
(i) Temporal modeling deficiency:
Visual features derived from fixed-size windows focus exclusively on localized segments, thus resulting in discontinuous semantic retrieval,
(ii) Modality gap:
Retrieved semantic features are fused with visual representations using simplistic operations (e.g., concatenation or basic attention mechanisms), which are inadequate for bridging the inherent divergence between visual and textual modalities, leading to inconsistencies that impair both localization accuracy and captioning quality.

To address these limitations, we propose that effective retrieval-augmented generation for dense video captioning involves exploiting the inherent temporal structure and rich semantic information embedded within video data. 
This is intuitively grounded in visual continuity: adjacent frames sharing similar visual and temporal contexts typically represent the same semantic event or action.
Inspired by this observation, we introduce explicit temporal-semantic modeling based on pseudo events to enhance contextual coherence and yield retrieved text semantics with temporal characteristics.
Furthermore, it is also essential to enhance visual representations using events rather than simply integrating frame-level or fragmented textual information.

\setlength{\belowcaptionskip}{-5pt} 

\begin{figure*}[h]
  \centering
  \includegraphics[width=0.92\textwidth]{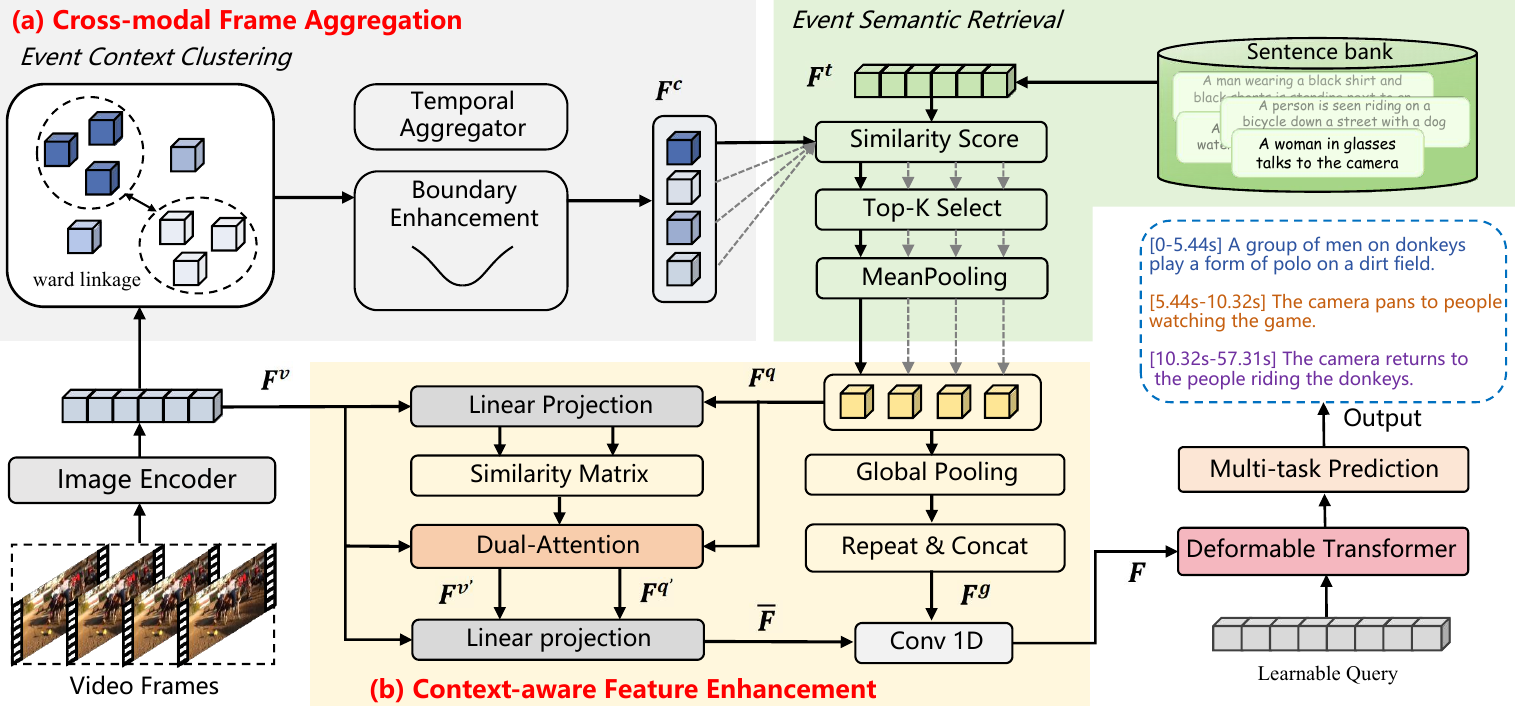}
  \caption{\textbf{The overview of our CACMI framework}. We employs a retrieval-augmented generation paradigm for DVC task. The pipeline begins with a pretrained CLIP image encoder extracting frame-level features. (a) Cross-modal Frame Aggregation (CFA). This module comprises two synergistic components: Event Context Clustering aggregates temporally and semantically consistent frame features to generate clustered event representations, and Event Semantic Retrieval matches relevant semantic information from a sentence bank via cosine similarity to produce retrieval-enhanced semantic features. (b) Context-aware Feature Enhancement (CFE). This module facilitates cross-modal interaction between retrieved textual features and visual representations, bridging the modality gap to generate enhanced frame features. Finally, a deformable transformer equipped with multi-task heads generates the joint outputs of event localization and captioning.}  
  \label{fig:overview}
\end{figure*}

As illustrated in Figure~\ref{fig:fig1}, we propose a novel framework called Context-aware Cross-Modal Interaction (CACMI), which leverages the explicit temporal-semantic structure in video data for the dense video captioning task.
First, we design a Cross-modal Frame Aggregation (CFA) module, which consists of two components: Event Context Clustering employs a temporal constraint mechanism to integrate visual features that share contextual and temporal coherence, and Event Semantic Retrieval performs cross-modal matching between the clustered event features and sentence bank, extracting event-aligned textual information.
Subsequently, a Context-aware Feature Enhancement (CFE) module is introduced to facilitate fine-grained integration of the visual features and retrieved textual features, enabling precise alignment of visual dynamics with linguistic semantics. Finally, we adopt a transformer encoder-decoder architecture with parallel multi-task heads to jointly perform event localization and caption generation. Our main contributions can be summarized as follows:
\begin{itemize}
\item We present CACMI, a novel dense video captioning framework featuring explicit temporal-semantic modeling, which leverages Context-Aware Cross-Modal Interaction to fully exploit the rich temporal structure and semantic information inherent in videos.

\item We introduce a Cross-modal Frame Aggregation module to extract temporally coherent, event-aligned semantic features through cross-modal retrieval, which is complemented by a Context-aware Feature Enhancement module that mitigates the visual-linguistic modality gap through query-guided feature refinement.

\item Extensive experiments on the ActivityNet Captions and YouCook2 datasets demonstrate the effectiveness of our model in dense video captioning task, highlighting its superior event localization capability.
\end{itemize}

\section{Related Work}

\subsection{Dense Video Captioning} Dense video captioning consists of two key subtasks: event localization and caption generation. Early methods typically followed a two-stage localization-description paradigm~\cite{krishna2017dense, wang2018bidirectional, wang2020event}, where salient temporal segments are first identified and then passed to language models for caption generation. However, this decoupled training strategy limits the mutual interaction between localization and captioning, hindering joint optimization. To overcome this, recent works adopt end-to-end frameworks that enable joint learning. PDVC~\cite{wang2021end} proposes a parallel decoding structure that shares intermediate representations across both subtasks. Vid2Seq~\cite{yang2023vid2seq} formulates dense video captioning as a sequence-to-sequence problem, leveraging transcribed speech from narrated videos as multimodal input and employing special text-time tokens to unify event detection and caption generation. CM$^{2}$~\cite{kim2024you} introduces a memory bank mechanism that preserves end-to-end learning while incorporating external textual knowledge to enhance caption quality.

Despite these advancements, achieving precise event localization and comprehensive semantic understanding remains a significant challenge, particularly without extensive pretraining on large-scale video datasets. We propose a novel DVC approach that integrates the inherent temporal representations of videos with retrieved textual semantics to improve both localization accuracy and caption generation.

\subsection{Retrieval-Augmented Captioning} Retrieval-augmented captioning enhances video understanding by integrating external textual knowledge, following the paradigm of Retrieval-Augmented Generation~\cite{jing2023memory}. In this setting, retrieved texts serve as supplementary context, enriching the original visual features and enabling deeper semantic comprehension. Recent DVC methods construct a text memory bank relevant to source videos, using retrieved sentences as multimodal input to boost both localization and captioning performance~~\cite{chen2023retrieval, ramos2023smallcap, kim2024you, kim2025hicm2}.

However, a key limitation persists: current retrieval strategies rely on manually designed sliding windows for search matching, neglecting contextual relationships across video segments. This approach results in fragmented retrieval that fails to capture the inherent temporal structure and coherent semantics of video content.
To overcome this, we propose a explicit temporal-semantic modeling framework for retrieval-augmented dense video captioning, which preserves the temporal structure of videos to ensure contextually consistent semantic retrieval.

\section{Method}

The goal of this study is to enhance event localization and caption generation from untrimmed videos by effectively leveraging the explicit temporal-semantic structure within video content. As illustrated in Figure~\ref{fig:overview}, we propose a novel framework named CACMI (Context-Aware Cross-Modal Interaction), which incorporates external textual knowledge from sentence bank and enhances visual features with pseudo-event semantics.
The model outputs a set of $N$ tuples ${(t^s_n, t^e_n, c_n)}^{N}_{n=1}$, where $N$ is the number of detected events in a video segment, $t^s_n$ and $t^e_n$ denote the start and end timestamps of the $n$-th event, and $c_n$ is the corresponding textual description.

\subsection{Cross-modal Frame Aggregation}
\subsubsection{Event Context Clustering.} 
We utilize pre-trained CLIP ViT-L/14~\cite{dosovitskiy2020image, radford2021learning} to extract frame-level visual features $\textbf{F}^v \in \mathbb{R}^{L \times d}$, where $L$ and $d$ are the number of clips and the feature dimension, respectively.
Within a video, frames associated with the same event often share similar background and foreground, leading to high similarity in their encoded representations. To capture the temporal-semantic correlations across frames, we apply agglomerative clustering to $\textbf{F}^v$ at the frame level.
Agglomerative clustering does not assume a fixed cluster shape, making it suitable for discovering flexible and diverse patterns in feature space. We adopt Euclidean distance as the similarity metric, which efficiently captures smooth variations in frame-level features. For cluster merging, we employ Ward linkage to minimizes the increase in within-cluster variance during hierarchical merging. This combination encourages the formation of compact and semantically coherent clusters with high intra-cluster similarity.

To further enhance temporal coherence, we incorporate a temporal aggregation constraint. Specifically, after clustering, we ensure that any two frames within a cluster are no more than $\textbf{t}_{\text{max}}$ apart in time. Frames that exceed this temporal threshold are assigned to a new cluster. This constraint guarantees that the resulting clusters maintain both semantic similarity and temporal continui. 
Finally, the video is segmented into $c$ clusters, each corresponding to a potential pseudo-event. The output is a set of cluster-level feature vectors $\textbf{F}^{c}=\{C_{i}\}_{i=1}^{c}$, where each $C_i$ represents the boundary-enhanced average of features within the $i$-th cluster.  Specifically, we generate a bell-shaped weight distribution centered at the cluster's midpoint and then invert and normalize these weights, thereby assigning higher importance to features near the cluster boundaries.

\subsubsection{Event Semantic Retrieval.}
To facilitate efficient cross-modal matching with an external textual corpus, we first preprocess the sentence bank using the CLIP text encoder, obtaining a set of textual feature embeddings $\textbf{F}^{t} \in \mathbb{R}^{M \times d}$, where $M$ is the total number of sentences in the corpus. We then compute a cosine similarity matrix $\mathbf{S} \in \mathbb{R}^{c \times M}$ between each pseudo-event visual feature and all text features:
\begin{gather}
\textbf{S} = \frac{\textbf{F}^c \, \textbf{F}^{t\top}} {{|| \textbf{F}^c||}\,{||\textbf{F}^{t}||}}
\end{gather}

Due to the large size of the sentence bank, we apply top-$k$ retrieval to each row of the similarity matrix $\mathbf{S}$, selecting the $k$ most relevant textual features for each pseudo-event. These retrieved features are aggregated into a condensed representation $\textbf{F}^{s} \in \mathbb{R}^{c \times k \times d}$:
\begin{gather}
\textbf{F}^{s} = \text{Top-K}(\mathbf{S})
\end{gather}

To form a unified representation for each event, we perform average pooling over the top-$k$ text features, resulting in the final retrieved semantic features $\textbf{F}^{q} \in \mathbb{R}^{c \times d}$:
\begin{gather}
\textbf{F}^{q} = \text{MeanPooling}(\textbf{F}^{s})
\end{gather}

\subsection{Context-aware Feature Enhancement }
As shown in Figure~\ref{fig:overview}, we introduce a fine-grained cross-modal fusion module to facilitate interactive refinement between retrieved textual features and visual representations. While CM$^{2}$~\cite{kim2024you} employs a transformer encoder-decoder architecture with shared self-attention weights for feature enhancement, this parameter-sharing scheme is insufficient for bridging the inherent semantic gap between visual and linguistic modalities. Visual features often contain substantial noise that is misaligned with textual context, and naive fusion methods such as simple addition and concatenation fail to capture fine-grained and semantically aligned visual information.

To address this challenge, we adopt a query-guided multimodal fusion module inspired by~\cite{xiong2016dynamic, sun2024tr}, which leverages textual queries to selectively suppress irrelevant visual elements and enhance semantically aligned regions, thereby enabling more accurate cross-modal alignment. The detailed structure of the module is shown in Figure~\ref{fig:overview}.
We begin by computing a similarity matrix between the frame-level visual features $\textbf{F}^v$ and the event-level textual queries $\textbf{F}^q$:
\begin{align}
    M = \frac{LP(\textbf{F}^v) \, LP(\textbf{F}^{q})^{\mathsf{T}}}{\sqrt{d}},
\end{align}
where $M \in \mathbb{R}^{L \times N}$ denotes the similarity matrix, and $LP$ represents the linear projection layer. Using $M$, we compute the dual-attention features: 
\begin{align}
    \textbf{F}^{v'} &= M_{col} \, \textbf{F}^{q}, \\
    \textbf{F}^{q'} &= M_{row} \, M_{col}^{\mathsf{T}} \, \textbf{F}^{v},
\end{align}
where $M_{\text{col}}$ and $M_{\text{row}}$ are the column-wise and row-wise softmax-normalized versions of $M$, respectively. Here, $\textbf{F}^{v'}$ captures event-level visual context guided by text, while $\textbf{F}^{q'}$ refines the query representation based on visual input.

Next, we concatenate the original frame features $\textbf{F}^{v}$ with the cross-attended features $\textbf{F}^{v'}$ and $\textbf{F}^{q'}$, followed by a linear projection to obtain the refined visual features $\overline{\textbf{F}}$:
\begin{align}
\overline{\textbf{F}} = LP ([\,{\textbf{F}^v} \vert \,\textbf{F}^{v'} \vert \,\textbf{F}^{q'} ]), 
\end{align}
where $\left[\cdot \vert \cdot \right]$ denotes feature concatenation. To incorporate global semantic guidance, we apply average pooling on $\textbf{F}^q$ to obtain a global text vector $\textbf{F}^g$, and replicate it across all frames to match the temporal dimension of $\textbf{F}^v$. Finally, we fuse this global context with $\overline{\textbf{F}}$ via channel-wise 1D convolution to produce the enhanced frame-level features:
\begin{align}
\textbf{F} = Conv_{1D} ([\,\textbf{F}^g \vert \,\overline{\textbf{F}} \,])
\end{align}

\subsection{Multi-task Prediction}
For event prediction, our CACMI framework incorporates a deformable transformer~\cite{zhu2020deformable} module followed by parallel multi-head predictors. The deformable transformer takes video features ${F}$ and a set of learnable queries $\{q_i\}^{N}_{i=1}$ as input, which produces semantic and temporal representations of events with encoder-decoder framework. Given the extracted event features, three separate prediction heads are utilized for dense video captioning process. 
\subsubsection{Localization Head.} 
The localization head consists of a multi-layer perceptron that predicts the temporal boundaries of events for each query. Specifically, it regresses the event center and temporal span, producing outputs in the form of tuples ${(t^s_i, t^e_i, c_i)}_{i=1}^{N}$, where $t^s_i$ and $t^e_i$ denote the predicted start and end times, and $c_i$ is the confidence score representing foreground probability. 
\subsubsection{Captioning Head.}
The backbone of the captioning head is an LSTM augmented with deformable soft attention around the predicted reference points~\cite{wang2021end}. At each decoding step $t$, the LSTM receives the context features $a_{i,t}$, the corresponding event query $q_i$, and the previous word $w_{i,t-1}$ to predict the next word $w_{i,t}$. This process continues until the full caption $S_i={w_{i,1},...,w_{i,S}}$ is generated for the $i$-th event, where $S$ denotes the caption length. 
\subsubsection{Event Counter.} The event counter is designed to predict the number of events in an input video. To achieve this, essential information from the event query ${Q}$ is first compressed using a max-pooling layer, followed by a fully-connected layer that outputs a fixed-size vector $f$. Each dimension of $f$ corresponds to a possible event count. During inference, the predicted number of events is given by $N = argmax(f)$.
Finally, we employ the Hungarian algorithm to match the predicted and ground-truth event tuples ${(t^s_n, t^e_n, c_n)}_{n=1}^{N}$, minimizing the matching loss defined as:
\begin{equation}
L_{\text{match}} = L_{\text{cls}} + \alpha L_{\text{loc}},
\end{equation}
where $L_{\text{cls}}$ is the focal classification loss, and $L_{\text{loc}}$ is the generalized IoU loss measuring the alignment between predicted and ground-truth temporal segments.

\subsection{Loss Function}
The overall training objective integrates four loss components: $L_{\text{cls}}$, $L_{\text{loc}}$, $L_{\text{count}}$, and $L_{\text{cap}}$. Specifically, $L_{\text{cls}}$ is the loss between predicted event classification and ground-truth labels, $L_{\text{loc}}$ is the generalized IoU loss for temporal boundary regression, $L_{\text{count}}$ is the cross-entropy loss for event count prediction, and $L_{\text{cap}}$ is the cross-entropy loss for word prediction across the generated captions. The final loss is a weighted sum of these components:
\begin{equation}
L = \alpha_{\text{cls}} L_{\text{cls}} + \alpha_{\text{loc}} L_{\text{loc}} + \alpha_{\text{count}} L_{\text{count}} + \alpha_{\text{cap}} L_{\text{cap}}
\end{equation}

\section{Experiments}
\subsection{Dataset}
We evaluate our CACMI on two widely-used benchmark datasets for dense video captioning: ActivityNet Captions~\cite{krishna2017dense} and YouCook2~\cite{zhou2018towards}. ActivityNet Captions comprises approximately 20,000 untrimmed YouTube videos, spanning over $700$ hours and encompassing a broad range of event categories such as sports, cooking, and social activities. Each video is annotated with an average of $3.7$ temporally localized captions, resulting in over $100,000$ sentence-level annotations with precise timestamps. Following the standard data split, we use $10,024$ videos for training, $4,926$ for validation, and $5,044$ for testing. 
YouCook2 focuses on instructional cooking videos and contains $2,000$ untrimmed YouTube videos, totaling $176$ hours of content. Each video is densely annotated with an average of $7.7$ captions, offering fine-grained temporal-textual alignments to support procedural video understanding. We follow the standard dataset split for training, validation, and testing purposes. It is worth noting that we only use videos that are still available on YouTube, which is $7\%$ fewer than in the original dataset.

\subsection{Implementation Details}
We sample video frames at a rate of 1 frame per second (FPS) for both datasets. To standardize input length, we either subsample or pad the frame sequences to a fixed number of frames $F$, where $F = 100$ for ActivityNet Captions and $F = 200$ for YouCook2. The number of event queries used in the Deformable Transformer is set to 10 for ActivityNet Captions and 100 for YouCook2 to account for the varying density of event annotations. For the Event Context Clustering module, we set the number of clusters to 10 and 20 for ActivityNet Captions and YouCook2, respectively. During Event Semantic Retrieval, we apply a soft top-$k$ selection with $k = 40$, allowing each pseudo-event center to retrieve the 40 most relevant text features from the memory bank. All remaining model hyperparameters are aligned with those used in CM$^{2}$~\cite{kim2024you}. All experiments are conducted using an NVIDIA RTX A6000 GPU.

\begin{table}[t]
    \begin{center}
        \footnotesize
        \setlength{\tabcolsep}{0.5pt} 
        \resizebox{\columnwidth}{!}{ 
            \begin{tabular}{l|>{\centering\arraybackslash}p{0.7cm}|>{\centering\arraybackslash}p{1.05cm}>{\centering\arraybackslash}p{1.05cm}>{\centering\arraybackslash}p{1.05cm}>{\centering\arraybackslash}p{1.05cm}}
                \toprule[1.5pt]
                Models &  PT  & B4↑ & M↑ & C↑ & S↑ \\
                \midrule[0.5pt]
                UEDVC (ECCV'22) & $\checkmark$ & - & - & - & 5.50 \\
                Vid2Seq (CVPR'23) & $\checkmark$ & - & 8.50 & 30.10 & 5.80 \\
                OmniVID (CVPR'24) \hspace{5pt} & $\checkmark$ & 1.73 & 7.54 & 26.00 & 5.60 \\
                \midrule[0.5pt]
                PDVC$^\dag$ (ICCV'21) & $\times$ & 2.21 & 8.06 & 29.97 & 5.92 \\
                CM$^{2\dag}$ (CVPR'24) & $\times$ & 2.38 & 8.55 & 33.01 &  \underline{6.18} \\
                E$^{2}$DVC (CVPR'25)  & $\times$ & \underline{2.43} &  \underline{8.57} &  \underline{33.63} & 6.13 \\
                \rowcolor{blue!20}
                CACMI (Ours) & $\times$ & \textbf{2.44} & \textbf{8.68} & \textbf{33.80} & \textbf{6.39} \\
                \bottomrule[1.5pt]
            \end{tabular}
        }
    \end{center}
    \caption{\textbf{Performance of Event Captioning in ActivityNet Captions}. B4, M, C, S denote BLEU4, METEOR, CIDEr and SODA\_c, respectively. $\dag$ indicates reproduced from official code. Bold means the highest score. Underline means 2nd score. PT denotes whether pretraining is conducted using additional video data.}
    \label{tab:anet_caption}
\end{table}

\begin{table}[t]
    \begin{center}
        \footnotesize
        \setlength{\tabcolsep}{0.5pt} 
        \resizebox{\columnwidth}{!}{ 
            \begin{tabular}{l|>{\centering\arraybackslash}p{0.7cm}|>{\centering\arraybackslash}p{1.05cm}>{\centering\arraybackslash}p{1.05cm}>{\centering\arraybackslash}p{1.05cm}>{\centering\arraybackslash}p{1.05cm}}
            \toprule[1.5pt]
            Models & PT& B4↑ & M↑ & C↑ & S↑ \\
            \midrule[0.5pt]
            Vid2Seq (CVPR'23) \hspace{5pt} & $\checkmark$ & - & 9.30 & 47.10 & 7.90 \\
            \midrule[0.5pt]
            PDVC$^\dag$ (ICCV'21) & $\times$ & 1.40 & 5.56 & 29.69 & 4.92 \\
            CM$^{2\dag}$ (CVPR'24) & $\times$ & 1.63 & 6.08 & 31.66 & 5.34 \\
            E$^{2}$DVC (CVPR'25) & $\times$ &  \underline{1.68} &  \underline{6.11} &  \underline{34.26} &  \underline{5.39} \\
            \rowcolor{blue!20}
            CACMI (Ours) & $\times$ & \textbf{1.70} & \textbf{6.21} & \textbf{34.83} & \textbf{5.57} \\
            \bottomrule[1.5pt]
        \end{tabular}
        }
    \end{center}
    \caption{\textbf{Performance of Event Captioning in YouCook2}. B4, M, C, S denote BLEU4, METEOR, CIDEr and SODA\_c, respectively. $\dag$ indicates reproduced from official code. Bold means the highest score. Underline means 2nd score. PT denotes whether pretraining is conducted using additional video data.}
    \label{tab:yc2_caption}
\end{table}

\begin{table}[t]
    \begin{center}
        \footnotesize
        \setlength{\tabcolsep}{0.1pt} 
        \renewcommand{\arraystretch}{1.2} 
        \resizebox{\columnwidth}{!}{ 
            \begin{tabular}{l|>{\centering\arraybackslash}p{0.5cm}|>{\centering\arraybackslash}p{1cm}>{\centering\arraybackslash}p{1cm}>{\centering\arraybackslash}p{1cm}|>{\centering\arraybackslash}p{1cm}>{\centering\arraybackslash}p{1cm}>{\centering\arraybackslash}p{1cm}}
                \toprule[1.5pt]
                \multirow{2}{*}{Models} & \multirow{2}{*}{PT} & \multicolumn{3}{c|}{ActivityNet Captions} & \multicolumn{3}{c}{YouCook2} \\
                & & \textit{F1}↑ & \textit{Rec.}↑ & \textit{Pre.}↑ & \textit{F1}↑ & \textit{Rec.}↑ & \textit{Pre.}↑ \\
                \midrule[0.5pt]
                Vid2Seq & $\checkmark$ & 53.29 & 52.70 & 53.90 & 27.84 & \textbf{27.90} & 27.80 \\
                \midrule[0.5pt]
                PDVC$^{\dagger}$ & $\times$ & 54.78 & 53.27 & 56.38 & 26.81 & 22.89 & 32.37 \\
                CM$^{2\dag}$ & $\times$ & 55.21 & 53.71 & 56.81 & 28.43 & 24.76 & 33.38 \\
                E$^{2}$DVC & $\times$ &  \underline{56.42} &  \underline{55.14} &  \underline{57.77} &  \underline{28.87} & 25.01 &  \underline{34.13} \\
                \rowcolor{blue!20}    
                CACMI (Ours) \hspace{2pt} & $\times$ & \textbf{57.10} & \textbf{55.89} & \textbf{58.05} & \textbf{29.34} &  \underline{25.54} & \textbf{34.63} \\
                \bottomrule[1.5pt]
            \end{tabular}
        }
    \end{center}
    \caption{\textbf{Performance of Event Localization in ActivityNet Captions and YouCook2 datasets.} Bold means the highest score. Underline means 2nd score. PT denotes pretraining from additional video datasets. $^{\dagger}$ denotes results reproduced from official implementation. All methods employ CLIP as the visual backbone. Rec. and Pre. denote average recall and average precision, respectively.}
    \label{tab:loc}
\end{table}

\begin{table}[t]
	\begin{center}
		\resizebox{1.0\columnwidth}{!}{
			\setlength{\tabcolsep}{1.61mm}{
				\begin{tabular}{cc|ccccc}
					\toprule[1.5pt]
					CFA & CFE &  BLEU4 & METEOR & CIDEr & SODA\_c & F1 \\ 
					\midrule[0.5pt]
					$\times$ & $\times$   & 2.38	      & 8.55          & 33.01  & 6.18  & 55.21 \\  
					$\checkmark$ & $\times$  & 2.37	      & 8.63 		  & 33.62 		  & 6.26  & 56.07 \\
					$\times$ & $\checkmark$   & 2.41	      & 8.59 		  & 33.48 		  & 6.31 & 56.95\\
					$\checkmark$ & $\checkmark$  & \textbf{2.44}	      &  \textbf{8.68} 		  & \textbf{33.80} 		  & \textbf{6.39} & \textbf{57.10} \\
					\bottomrule[1.5pt]
				\end{tabular}
		}
	}
	\end{center}
	\caption{\textbf{Performance of different components}. CFA and CFE denote cross-modal frame aggregation and context-aware feature enhancement, respectively.}
	\label{tab:components}
\end{table}

\begin{table}[t]
	\begin{center}
		\resizebox{1.0\columnwidth}{!}{
			\setlength{\tabcolsep}{1.61mm}{
				\begin{tabular}{c|ccccc}
					\toprule[1.5pt]
					$N_{\text{cluster}}$ &  BLEU4 & METEOR & CIDEr & SODA$_c$ & F1 \\ 
					\midrule[0.5pt]
					3   & 2.32	      & 8.53          & 32.84         & 6.12  & 54.91 \\  
					5  & 2.35	      & 8.58 		  & 33.15 		  & 6.21  & 54.87 \\
					7   & 2.36	      & 8.63 		  & 33.24 		  & 6.28 & 56.02 \\
					\textbf{10}       & \textbf{2.44}	      &  \textbf{8.68} & \textbf{33.80} 		  & \textbf{6.39}   & \textbf{57.10} \\
					15  & 2.28	      & 8.49 		  & 32.98 		  & 6.19 & 55.15 \\
					$\gamma$  & 2.34	      & 8.60 		  & 33.43 		  & 6.29 & 56.23 \\
					\bottomrule[1.5pt]
				\end{tabular}
		}
	}
	\end{center}
	\caption{\textbf{Ablation study on the number of event clusters.} We report the results on the ActivityNet Captions. The best performance is highlighted.}
	\label{tab:abla_cluster}
\end{table}

\begin{table}[t]
	\begin{center}
		\resizebox{1.0\columnwidth}{!}{
			\setlength{\tabcolsep}{1.61mm}{
				\begin{tabular}{>{\centering\arraybackslash}p{1.5cm}|ccccc}
					\toprule[1.5pt]
					$top_k$ &  BLEU4 & METEOR & CIDEr & SODA\_c & F1 \\ 
					\midrule[0.5pt]
					10  & 2.23	      & 8.49 		  & 32.20 		  & 6.32 & 55.95 \\
					20  & 2.31	      & 8.60 		  & 32.26 		  & 6.21 & 55.50 \\ 
					\textbf{40}  & \textbf{2.44}	      & \textbf{8.68} 		  & \textbf{33.80} 		  & \textbf{6.39} & \textbf{57.10} \\
					60  & 2.27	      & 8.50 		  & 32.25 		  & 6.25 & 56.39 \\
					80  & 2.25	      & 8.53		  & 32.57 		  & 6.27 & 56.15 \\

					\bottomrule[1.5pt]
				\end{tabular}
		}
	}
	\end{center}
	\caption{\textbf{Ablation study on the top-k selection number for retrieval.} We report the results on the ActivityNet Captions. The best performance is highlighted.}
	\label{tab:abla_topk}
\end{table}

\subsection{Evaluation Metrics}

The evaluation of our model is conducted from two complementary perspectives: (i) Event Captioning Performance: To assess the quality of generated captions, we utilize standard metrics including CIDEr~\cite{vedantam2015cider}, which computes TF-IDF weighted n-gram consensus, BLEU4~\cite{papineni2002bleu}, which measures 1 to 4 gram precision, and METEOR~\cite{banerjee2005meteor}, which incorporates synonym matching and word order alignment. These scores are averaged across multiple IoU thresholds \{0.3, 0.5, 0.7, 0.9\} to ensure robustness. Additionally, we report SODA$\_\text{c}$~\cite{fujita2020soda}, a metric designed to evaluate storytelling ability and the coherence of the overall caption sequence. (ii) Event Localization Performance: To measure localization accuracy, we report the average precision, average recall, and F1 scores, each calculated across IoU thresholds \{0.3, 0.5, 0.7, 0.9\}, providing a comprehensive view of temporal alignment performance.

\begin{figure*}[h]
  \centering
  \includegraphics[width=1.0\textwidth]{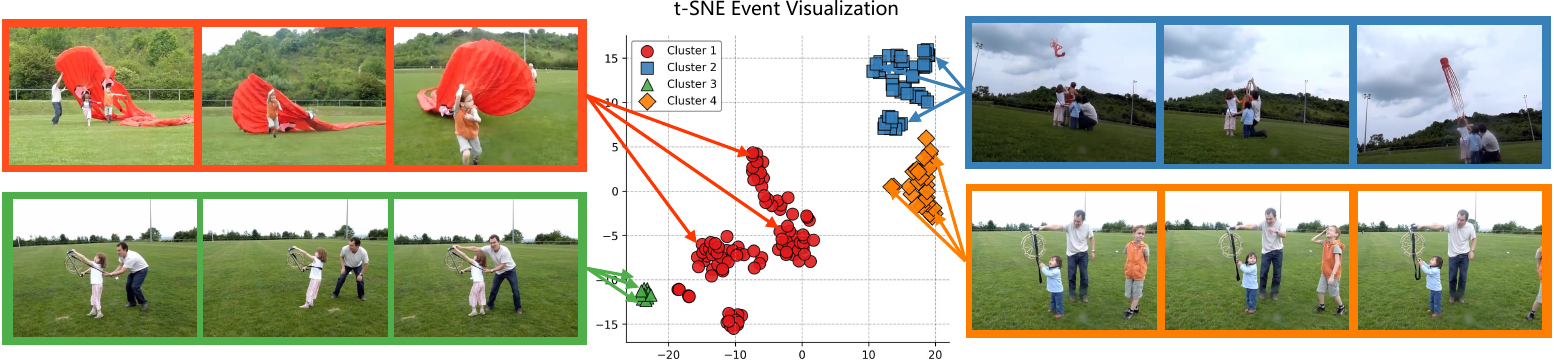}
  \caption{Visualization of event features. The t-SNE projection illustrates a two-dimensional embedding space, where grouped points within the same cluster indicate temporal correlation and semantic similarity. This demonstrates that the frame aggregation module effectively constructs discriminative event representations while preserving meaningful temporal information. }
\label{fig:tsne}
\end{figure*}

\subsection{Results} 
\subsubsection{Dense Video Captioning Performance.} In Table~\ref{tab:anet_caption} and Table~\ref{tab:yc2_caption}, we compare our CACMI framework with the state-of-the-art methods on the ActivityNet Captions and YouCook2 datasets. As shown in Table~\ref{tab:anet_caption}, our CACMI consistently outperforms the strong baseline CM$^{2}$~\cite{kim2024you} across all four metrics CIDEr, METEOR, BLEU4, and SODA\_c and achieves highly competitive results with the state-of-the-art E$^{2}$DVC~\cite{wu2025event}. Remarkably, our method even surpasses some pretrained models that leverage large-scale external video data, highlighting the effectiveness of our temporal-semantic modeling.
For Table~\ref{tab:yc2_caption}, Vid2Seq~\cite{yang2023vid2seq} achieves higher scores than our method on YouCook2 dataset. The performance gap is attributed to the limited training videos coupled with highly diverse event semantics for YouCook2 dataset, where Vid2Seq benefits from broader domain coverage.
Importantly, our CACMI significantly outperforms all non-pretrained baselines in SODA$\_\text{c}$, a metric designed to evaluate the storytelling quality and coherence of multi-event captions. This result underscores our model’s ability to effectively capture temporal and semantic dependencies within untrimmed video streams, leading to more contextually rich and consistent caption generation. 
\subsubsection{Event Localization Performance.}
We further evaluate the event localization capability of our CACMI. As shown in Table~\ref{tab:loc}, our CACMI achieves the state-of-the-art performance on both benchmark datasets. On the ActivityNet Captions dataset, our CACMI attains an F1 score of 57.10, with a recall of 55.89 and precision of 58.05, indicating a strong balance between accurate detection and coverage of relevant events. On the more challenging YouCook2 dataset, our method achieves an F1 score of 29.34, with 25.54 recall and 34.63 precision, demonstrating robustness in complex, fine-grained scenarios.
These results highlight the effectiveness of our temporal-semantic retrieval framework, which not only improves captioning quality by modeling temporally coherent semantics, but also strengthens localization accuracy by incorporating rich temporal cues into the event boundary prediction process.

\subsection{Ablation Studies}
\begin{figure*}[h]
  \centering
  \includegraphics[width=1\textwidth]{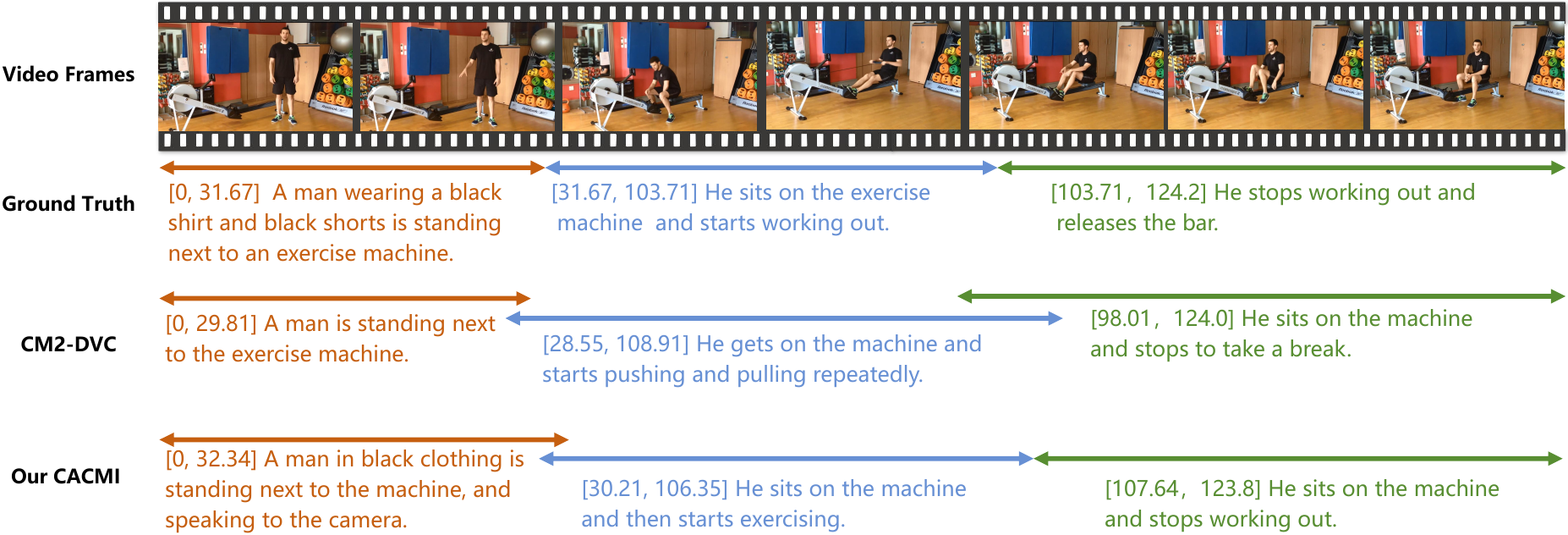}
  \caption{\textbf{Visualizations of dense event captioning prediction on ActivityNet Captions.} We present the results of the ground truth, the baseline CM$^{2}$ and our method.}
  \label{fig:pred}
\end{figure*}

\subsubsection{Analysis of Different Components.} Table~\ref{tab:components} presents ablation results on the ActivityNet Captions dataset, evaluating the contributions of the key components in our framework: Cross-modal Frame Aggregation (CFA) and Context-aware Feature Enhancement (CFE).
The CFA module captures temporally consistent event information and enhances visual representations by retrieving semantically relevant text features, while the CFE module aligns visual dynamics with textual semantics through a context-query attention mechanism.
From the results, CFE alone substantially boosts both event localization and caption generation, particularly improving temporal boundary precision and enhancing narrative coherence. Although CFA independently contributes noticeable gains in both tasks, the combination of CFA and CFE yields the best overall performance, demonstrating strong synergistic effects. This integration enables CACMI to effectively retain fine-grained temporal structure, leading to superior dense video captioning performance.

\subsubsection{Effect of the Number of Event Clusters.} As shown in Table~\ref{tab:abla_cluster}, we investigate the impact of the number of event clusters used during temporal clustering. The clustering process is guided by a hyperparameter $\gamma$, which adaptively determines the optimal number of clusters from 5 to 10. 
When the number of clusters is too small, the model captures only coarse event priors, limiting fine-grained temporal representation and resulting in marginal performance. Conversely, too many clusters lead to over-segmentation and noise from fragmented boundaries. Our results show that optimal performance is achieved with $N_{cluster}$ is 10, balancing temporal coherence and semantic granularity.

\subsubsection{Effect of the Number of Retrieved Sentences.} 
Table~\ref{tab:abla_topk} shows the impact of varying the number of retrieved textual features on model performance. The number of clustered events is fixed at 10, then we evaluate the influence of different top-k values during the cross-modal retrieval process. When the top-k value is set too low, each cluster has access to only a limited number of supplemental sentences. This restricted retrieval fails to provide sufficient semantic diversity, limiting the model's ability to capture rich contextual cues. Conversely, an excessively high top-k value introduces redundant or less relevant sentences, which can dilute the informative content and obscure key semantic signals, ultimately hindering the model’s performance.
Empirically, the model achieves optimal performance when the top-k value is set to 40, striking an effective balance between contextual richness and semantic relevance.

\subsection{Qualitative Comparison}
\subsubsection{Visualization of Event Clusters.}
Figure~\ref{fig:tsne} shows the frame-level features aggregated by the Event Context Clustering module, providing qualitative evidence of the temporal coherence within the generated event representations. We first apply PCA to reduce the dimensionality of the high-dimensional visual features, followed by t-SNE projection into a two-dimensional embedding space for visualization. 
The results show that features grouped within the same cluster correspond to video segments exhibiting strong temporal continuity and semantic similarity, while features from different clusters represent visually distinct content.
\subsubsection{Visualization of Predicted Results.}
Figure~\ref{fig:pred} illustrates a qualitative example of event predictions generated by our CACMI. Compared to the baseline CM$^{2}$, our CACMI produces more precise event boundaries, demonstrating significantly enhanced localization capabilities. Simultaneously, it captures richer semantic details, enabling more accurate and contextually relevant descriptions of video events. By leveraging contextual semantics, our CACMI achieves a deeper understanding of video content while maintaining robust event localization performance. 

\section{Conclusion}
We propose Context-Aware Cross-Modal Interaction (CACMI), a novel framework for explicit temporal-semantic modeling in the dense video captioning task. Our CACMI effectively leverages the temporal dependencies within video content and the semantic knowledge embedded in text corpus through a unified cross-modal interaction strategy.
The framework consists of two core components: Cross-modal Frame Aggregation, which enhances contextual understanding and semantic enrichment by grouping temporally coherent frames and retrieving relevant text features; and Context-aware Feature Enhancement, which bridges the semantic gap between modalities by aligning visual dynamics with textual semantics.
Comprehensive experiments on the ActivityNet Captions and YouCook2 datasets demonstrate that our CACMI achieves the sota results, significantly improving event localization accuracy.

\section{Acknowledgments}
This work was supported by the National Natural Science Foundation of China (Grant Nos.\ U21A20515, 62376271, U22B2034, 62171321, 62572059, and 62365014), the Beijing Natural Science Foundation (Grant Nos.\ L231013, L241056), the Shenzhen Science and Technology Program (Grant No.\ CJGJZD20240729141906008), the Open Project of the Key Laboratory of Computing Power Network and Information Security (Grant No.\ 2024PY021), and the Open Project Program of the State Key Laboratory of Virtual Reality Technology and Systems, Beihang University (Grant No.\ VRLAB2025B03).

\bibliography{aaai2026}

\end{document}